# Can rationality be measured?

Tshilidzi Marwala

University of Johannesburg

South Africa

Email: tmarwala@gmail.com

**Abstract**

This paper studies whether rationality can be computed. Rationality is defined as the use of complete information, which is processed with a perfect biological or physical brain, in an optimized fashion. To compute rationality one needs to quantify how complete is the information, how perfect is the physical or biological brain and how optimized is the entire decision making system. Rationality of a model (i.e. physical or biological brain) is measured by the expected accuracy of the model. Rationality of the optimization procedure is measured as the ratio of the achieved objective (i.e. utility) to the global objective. The overall rationality of a decision is measured as the product of the rationality of the model and the rationality of the optimization procedure. The conclusion reached is that rationality can be computed for convex optimization problems.

**Introduction**

Rationality is one of those elusive concepts that philosophers have invented. Simply put, to be rational is to be logical (Anand, 1993; Marwala, 2018a). But this linguistic explanation of rationality is inadequate. The more complete definition of being rational is to analyze or reach a conclusion (e.g. decision) by using all the information available and processing such information in an optimized manner. The challenge is how about if the information is incomplete? What if the information processing engine is less than perfect? How about if it is not possible to process such information? What is the difference between the biological and physical brains? The biological brain is the natural brain such as a human brain whereas the physical brain is the artificial brain such as artificial neural network, which is a type of artificial intelligence (AI). Herbert Simon coined these limitations of information, brain and optimization *bounded rationality* (Simon, 1982). Bounded rationality simply means limited rationality. If rationality can be limited, can we compute the extent of this limitation? Marwala (2018a) asked whether the physical brain, such as AI, is rational and concluded that even though the physical brain is not rational, advances in technology developments make it more rational, and that it is more rational than the biological brain. Comparing physical to biological brain evokes the question of whether it is possible to compute or measure rationality. In order to answer the question of the quantification of rationality, it is important to break rationality into its three components, and these are information, information processing engine (physical or biological brain) and optimization. Then we can investigate if rationality of each of these components can be computed. Then if rationality of these components can be computed, how these quantifications can be aggregated to form a composite estimation of rationality. This approach of looking at the components of decision making nodes and evaluating their rationality is called the procedural way of evaluating rationality. The other way of calculating rationality is the substantive method, where the outcome of the decision making process is evaluated and based on its performance, the extent of rationality is estimated. Herbert Simon describes these as procedural and substantive rationality (Simon, 1976).

**Information**

Rational decisions are based on relevant information. If one uses irrelevant information to make a decision then one is being irrational. Superstition is the practice of using irrelevant information to make a decision or draw a conclusion. For example, there was a belief that if one encounters a black cat crossing the road then this was an indication of bad luck (Marwala, 2014). The other superstition is that if one hears an owl hooting then one will have death in his/her family. Of course neither does the hoot of an owl has anything to do with death nor does the crossing of the black cat has anything to do with bad luck. All these are conclusions that are derived based on irrelevant information and thus are irrational. It turns out that human beings are not rational beings (Kahneman, 2011). Ruling out rationality based on irrelevant information is easy. There are two reasons why relevant information can truncate rationality and these are either the information is incomplete or it is imperfect (Marwala, 2009). Complete information is very difficult to obtain but can be accessed for very simple problems. For some classes of problems, incomplete information can be estimated. Artificial intelligence has been successfully used to estimate missing data (Marwala, 2009; Marwala, 2019). AI is a computational technique, which is used to make machines intelligent, and has been successfully used in engineering problems (Marwala, 2010, 2012 & 2018b), to predict interstate conflict (Marwala and Lagazio, 2011), to model financial markets (Marwala, 2013; Marwala and Hurwitz, 2017) and to model rationality (Marwala, 2014 & 2015). The difficulty with imperfect information is that very often one cannot determine how imperfect the information is. As a result it is difficult to establish the impact of imperfect information on the rationality of a decision that is taken.

**Model: Biological and Physical Brain**

To quantify how rational the decision is, relevance, degrees of completeness and perfection of information as well as the effectiveness of a decision making machine, whether a biological or physical brain, are taken into account. This is a difficult route to take and thus will be difficult to quantify rationality. The other way of quantifying rationality is by evaluating the effectiveness of the decision machine made. The information that is collected is processed by some device in order to make a rational decision. The idea of predicting the future in decision making is important otherwise the decision will be random. For humans such a device is the brain and for machines it is based on artificial intelligence. The human brain takes the input information and process it. It has been observed that humans are irrational beings and, therefore, it is not easy to quantify their degree of rationality (accuracy) in advance. For AI machines, they are trained and validated using statistical procedures and one is able to estimate their expected effectiveness from the validation data. This accuracy level is linked to the rationality of the AI machine. The more accurate the AI machine, the more rational it is. There are methods that have been advanced to improve the accuracy or rationality of the AI machines. One of these is that the model shouldn't be complex. This is what is called the Occam's razor and is named after William of Occam (Adams, 1987; Keele, 2010). Occam's razor simply says that the simplest model that explains all observations is the one that has the highest probability of being correct. Of course the correctness of the model is a contentious issue because George Box famously declared that "all models are wrong" (Box, 1976). What he meant was that models are nothing but estimation of reality and they are not reality. If we are to take George Box's conclusion to its rational conclusion then all models are not fully rational, which is the conclusion that was reached by Herbert Simon. In essence George Box's 'all models are wrong" is a statistical version of Herbert Simon's theory of bounded rationality. The degree of rationality of models given the information that is fed into these models can be estimated by how accurate these models are.

**What about the Uncertainty Principle?**

At the turn of the previous century Max Planck came up with the quantum theory (Planck, 1900). On trying to mathematically model the black body radiation problem, he realized that if he assumed that energy was in the form of packets called quanta then the model is able to predict measurements accurately. At first he thought it was just a mathematical trick. Albert Einstein used this mathematical trick to explain the photoelectric effect and this is the work for which he was awarded the Nobel Prize (Einstein, 1905). Then it became clear that the quantum theory is not just a mathematical trick that explains the black body radiation problem but it also explains the photoelectric effect. Later on Werner Heisenberg observed that because of the theory of quantum mechanics one cannot know the position and the velocity of an electron at the same time (Heisenberg, 1927). If one knows the velocity of the electron then one does not know its position and vice versa. This is what is called the Heisenberg uncertainty principle and it is not because of the effects of measurements but it is the fundamental law of quantum mechanics. Albert Einstein was so impressed with the Heisenberg uncertainty principle that he nominated Heisenberg for a Nobel Prize. A similar concept exists in the interplay between the model accuracy and the number of input variables. If one uses too few variables to create a model, then the model will not be rational, i.e. accurate, enough. If one uses too many variables to create the model, the model will be too complex and because of Occam's razor principle, it will be too complex, thereby, compromising the rationality of the model (Franklin, 2001). The concept of using too many variables in the model is called the *curse of dimensionality* (Bellman, 1957 & 1961). This interplay between rationality, model complexity and the curse of dimensionality is one of the reasons why models are always rationally bounded.

**Optimization**

Given the fact that we have related accuracy of models to rationality another element of rationality is whether the process of using information to build the models is efficient. One way of achieving that efficiency is through optimization. Why is optimization important in rationality? Suppose Denga wants to fly from New York to Los Angeles and she decides to fly from New York to Tokyo to Los Angeles for no other reason except that she wanted to move from New York to Los Angeles. We will characterizing this move as irrational. Travelling from New York to Los Angeles directly is more rational than travelling from New York to Tokyo to Los Angeles *ceteris paribus*[1]. The direct New York to Los Angeles trip is more optimized than the New York-Tokyo-Los Angeles trip. So optimization is important for rationality. When AI machines are used, the optimization process includes two aspects and these are computational efficiency and model accuracy. To optimize both processing effort and model accuracy is known as multi-objective optimization (Miettinen, 1999). In multi-objective optimization the weights given to processing effort and model accuracy are subjective. For each weights given to these objective functions, i.e. processing effort and model accuracy, there is an optimal solution. The combination of these optimal solutions, given different weights of processing effort and model accuracy, forms the Pareto optimal frontier (Pareto, 1897; Mathur, 1991). There is no solution on the Pareto optimal frontier which is the true maximum solution. The second difficulty with optimization is that it is not clear whether an optimal solution is the global optimum solution. In the theory of optimization there are two types of problems and these are convex and non-convex problems (Borwein and Lewis, 2000). Convex problems are those where one is guaranteed to identify a global optimum solution whereas the non-convex problems are those where one is not guaranteed an optimal solution. Now how does one quantify rationality given information, model and optimization process as well as the multi-objective nature of the problem, which is subjective? Does this imply that there is an element of subjectivity in the quantification of rationality? For convex problems, the accuracy of the model is adequate to quantify rationality. For non-convex problems rationality is subjective. The way to quantify rationality in the optimization process is to estimate the ratio between the achieved objective and the global objective. This means if we cannot identify the global optimum objective then it is not possible to quantify the rationality of the optimization procedure. This means only convex problems, where the global objective is identifiable, can we be able to quantify rationality.

---
[1] Other things equal

**Classification of Rationality**

One of the steps that has been taken is the idea of classifying the extent of rationality. In the financial markets customers exchange goods and services and the price of this exchange is driven by the laws of demand and supply (Stringham and Curott, 2015). One such markets is the stock exchange where investors buy stocks from companies at prices determined by the laws of demand and supply. Nobel Laureate Eugene Fama studied market efficiency and came up with the efficient market hypothesis (Fama, 1965). The efficient market hypothesis claims that markets are efficient but this proposition has been proven not to be universally correct. There is a relationship between efficiency and rationality. Efficient markets are rational and inefficient markets are not fully rational. In financial sciences, there are three ways of classifying efficiency of markets and these are strong, semi-strong and weak forms of market efficiency (Fama, 1996 & 2008). The weak form of market efficiency uses publicly available information to price the financial assets. The semi-strong market efficiency claims that prices reflect publicly available information and are able to quickly adapt to new information. The strong market efficiency claims that prices reflect both hidden and publicly available information. If we use the linkage between rationality and efficiency, then the strong market efficiency implies strong rational markets, semi-strong efficiency implies semi-strong market rationality and weak market efficiency implies weak rational markets. This of course is the quantification of rationality of the markets. Market efficiency assumes what is called the theory of rational expectations, which prescribes that agents reflect all the information that is available and, therefore, cannot be consistently wrong on predicting the future, a task that AI models are good at (Muth, 1961). This way of classifying rationality based on the efficient market hypothesis implies that only information and agents that predict the future are important in determining market rationality. However, we have observed that the uncertainty principle between information and prediction agents (i.e. the model) as well as the optimization of the whole functioning of the markets are important in the quantification of rationality of the markets.

**Rationality Quantification**

As described above rational decision making uses complete information and a model to make an optimized decision. Rationality of a decision is based on rationality of the model as measured by its accuracy as well as rationality of the optimization procedure, which is quantified by the ratio of the achieved objective to the global objective. The global objective is only attainable for convex problems. The overall rationality of a decision can be calculated as the product of the rationality of the model and the rationality of the optimization procedure. This, therefore, means that rationality can be computed provided the problem in question is a convex problem.

**Conclusion**

This paper attempted to answer the question on whether rationality can be quantified. On answering this question we considered a model which takes information and makes a decision. We concluded that the measure of the rationality of the model depends on how accurate the model is. The accuracy of this model depends on the completeness and the degree of perfection of the information as well as the uncertainty principle interplay between the model and information. There is an added dimension to the rationality of the process, which is that the whole process should be globally optimized. This situation can only be guaranteed for convex problems. The rationality of the optimization procedure is measured as the ratio of the achieved objective to the global objective. The overall rationality of a decision is measured as the product of the rationality of the model and the rationality of the optimization procedure. The conclusion is, therefore, that rationality can be computed for convex optimization problems.